\documentclass[a4paper,fleqn]{cas-sc}

\usepackage[numbers]{natbib}
\def\tsc#1{\csdef{#1}{\textsc{\lowercase{#1}}\xspace}}
\tsc{WGM}
\tsc{QE}
\tsc{EP}
\tsc{PMS}
\tsc{BEC}
\tsc{DE}
\begin{document}
\let\WriteBookmarks\relax
\def\floatpagepagefraction{1}
\def\textpagefraction{.001}
\shorttitle{A Comprehensive Image Dataset to Classify Diseased and Healthy Mango Leaves}
\shortauthors{S. I. Ahmed et~al.}
%\begin{frontmatter}

\title [mode = title]{MangoLeafBD: A Comprehensive Image Dataset to Classify Diseased and Healthy Mango Leaves}

\author[1]{Sarder Iftekhar Ahmed}
\address[1]{Dept. of Computer Science and Engineering, East West University, Dhaka, Bangladesh}
\ead{iftekhar.sarder@gmail.com}
\author[2]{Muhammad Ibrahim} [orcid=0000-0003-3284-8535]
\address[2]{Dept. of Computer Science and Engineering, Univeristy of Dhaka, Dhaka, Bangladesh}
\ead{ibrahim313@du.ac.bd}
\author[1]{Md. Nadim}
\ead{nadimewu.cse@gmail.com}
\author[1]{Md. Mizanur Rahman}
\ead{mizanurewu@gmail.com
}
\author[1]{Maria Mehjabin Shejunti}
\ead{mariamehjabin10@gmail.com
}
\author[1]{Taskeed Jabid}
%\address[1]{Dept. of Computer Science and Engineering, Ahsanullah University of Science and Technology, Dhaka, Bangladesh}
\ead{taskeed@ewubd.edu}
\author[1]{Md. Sawkat Ali*}
\cortext[1]{Corresponding author}
\ead{alim@ewubd.edu}

\begin{abstract}
Agriculture is of one of the few remaining sectors that is yet to receive proper attention from machine learning community. The importance of dataset in machine learning discipline cannot be overemphasized. The lack of standard and publicly available datasets related to agriculture impedes  practitioners of this discipline to harness full benefit of these powerful computational predictive tools and techniques. To improve this scenario, we develop, to the best of our knowledge, the first-ever standard, ready-to-use, and publicly available dataset of mango leaves. The images are collected from four mango orchards of Bangladesh, one of the top mango-growing countries of the world. The dataset contains 4000 images of about 1800 distinct leaves covering seven diseases. We also report accuracy metrics, namely precision, recall and F1 score of three deep learning models. Although the dataset is developed using mango leaves of Bangladesh only, since we deal with diseases that are common across many countries, this dataset is likely to be applicable to identify mango diseases in other countries as well, thereby boosting mango yield. This dataset is expected to draw wide attention from machine learning researchers and practitioners in the field of automated agriculture.

%PREVIOUS ABSTRACT: Mango is the fruit in the world which have huge socioeconomic implications. Various diseases and pests destroy mango leaves, causing socioeconomic loss. There has been research done on diseases of mango leaves in the past, but it does not provide a complete solution for identifying common mango leaf diseases.  For this study, the major objective is to find a solution to the problem of correctly diagnosing a wide range of frequent mango leaf illnesses. DeepCONVSVM, a hybrid of CNN and SVM, has been proposed as a convolutional neural network based architecture to accomplish this. CNN and ResNet50 have been built to see if the proposed model is efficient. According to the study's experimental data, the proposed model outscored the other two models with 91\% accuracy, 91\% precision, and 91\% recall.
\end{abstract}

\begin{comment}
\begin{graphicalabstract}
\includegraphics{figs/grabs.pdf}
\end{graphicalabstract}

\begin{highlights}
\item Research highlights item 1
\item Research highlights item 2
\item Research highlights item 3
\end{highlights}

\end{comment}


\maketitle

\section{Introduction}
Food and Agricultural Organization (FAO) reports in a survey that the food production must increase by 70\% by 2050 to meet the need of growing population of the world\footnote{\url{https://www.fao.org/news/story/en/item/35571/icode/}}. One of the main barriers to increased food production is the diseases of the plants. While traditionally plant diseases are identified by farmers and crop experts by visually examining the plants, in recent times the rapid surge of technological innovations has paved the way to use technologies to assist the humans in effective diagnosis of various plant diseases. Technology has started to change the disease identification and treatment. These relatively cheap apparatuses,  instruments, and techniques can be leveraged to cover large and geographically sparse crop fields in relatively less time and cost.

Among the cutting-edge technologies of today, machine learning discipline stands high as these techniques can predict future events at an astonishing level, provided sufficient and appropriate past data are available to them. Machine learning is a multi-disciplinary applied science at the core of which lie mathematics and statistics. Harnessing the low-cost computer hardware, these algorithms are being successfully applied in a range of human sectors. Since agriculture is one of the major sectors of a country, it is no surprise that machine learning easily finds its application here \cite{ibrahim_benos2021machine}, \cite{ibrahim_sharma2020systematic}. Indeed, this discipline has been changing the traditional way of diagnosing and treating various plant diseases \cite{ibrahim_yang2017machine}.

Key to the successful use of machine learning is having a good dataset to begin with. These algorithms extract hidden patterns from the dataset (which is called building or training the model), and then based on these learnt patterns they predict future events. So there is a high correlation between the quality of the dataset and the performance of a machine learning system.  The quality of a dataset can be weighed in terms of its size, intra-class integrity, inter-class dissimilarity, quality of labels i.e. presence of noise in the labels, among other things. A dataset must be a true representative of the real-life scenario where the learning system is going to be applied, otherwise the power of these models cannot be utilized.\footnote{In technical terms, the distribution of the training set and test set should be similar. This is one of the major foundations of machine learning.} So different datasets are needed for different prediction tasks, and this makes the datasets, in and of itself, valuable.

\subsection{Motivation}
A very promising area of research and innovation is the application of machine learning models for detecting plant diseases from leaf images \cite{ibrahim_mohanty2016using}. So the researchers in this field need readily-available representative datasets to develop effective machine learning systems. However, this type of real-life datasets for plant disease detection is not plentiful. We survey the existing literature  and found very few such datasets. The most popular and used dataset is named PlantVillage\footnote{\url{https://www.kaggle.com/datasets/emmarex/plantdisease}. Note that by using augmentation techniques, this dataset is expanded containing 87000 images here: \url{https://www.kaggle.com/datasets/vipoooool/new-plant-diseases-dataset}. Since this is not any new dataset, we omit its discussion.}. PlantVillage contains images of three plant leaves: pepper (1 disease and 1 healthy categories), potato (1 disease and 1 healthy categories), and tomato (9 diseases and 1 healthy  categories). The other dataset we found is DigiPathos\footnote{\url{https://www.digipathos-rep.cnptia.embrapa.br/}} \cite{ibrahim_barbedo2018annotated} that contains 2326 images of 21 plant leaves of Brazil. However, this dataset is not widely known to the researchers as yet. %No report on which countries data, so it misses a very important information.

In this research we aim to develop a standard, ready-to-use, and publicly available dataset of images of mango leaves of Bangladesh. Mango is one of the most popular fruits in the world (specifically, the 5th or 6th most popular food in terms of production with a yearly yield of about 50 million metric tons \cite{ibrahim_mango1, ibrahim_mango2}). However, few counties are eligible for growing good quality mangoes\footnote{The top 10 mango producing countries produce almost 40 out of approximately 50 million metric ton yearly yield \cite{ibrahim_mango3}}. Being the 9th highest mango-producing country in the world \cite{ibrahim_mango3}, Bangladesh has a huge potential for increasing its mango production by utilizing state-of-the-art machine learning technologies. However, we have not found a single dataset on mango leaves of Bangladesh which impedes the researchers to apply machine learning models in this nascent but promising domain.

We found only two datasets in the  existing literature that have even distant resemblance with our intended dataset. The first one is Indian mango leaves\footnote{\url{https://www.tensorflow.org/datasets/catalog/plant_leaves} contains the description and other resources regarding this dataset, and the original dataset can be found here: \url{https://data.mendeley.com/datasets/hb74ynkjcn/1}}, collected in India that contains 12 plant leaves, namely Mango, Arjun, Alstonia Scholaris, Guava, Bael, Jamun, Jatropha, Pongamia Pinnata, Basil, Pomegranate, Lemon, and Chinar. These plants are divided into only two categories: diseased and healthy, so there is no way to deal with multiple diseases. The second one is Indonesian mango leaves\footnote{\url{https://data.mendeley.com/datasets/94jf97jzc8/1}} that contains around only 510 mango leaf images\footnote{The dataset developers later use augmentation techniques to expand these 510 images to 62000 images.} that are affected by various pests. However, none of these two datasets can be considered as standard because of their very smaller size and less number of disease categories covered.

We summarize in Table~\ref{tab:datasets} the key properties of the datasets discussed above.
\begin{table}[width=.9\linewidth,cols=4,pos=h]
\caption{Existing relevant datasets }
\label{tab:datasets}
\begin{tabular*}{\tblwidth}{@{} LLLL@{} }
\toprule
Dataset & Plant Names & No. of Images\\
\midrule
 PlantVillage  & Pepper, potato, and tomato & 54000  \\
 DigiPathos & Citrus, soybean etc. & 2326 \\
 Indian leaves & Mainly medicinal plants (Arjun, Basil etc.) & 4503 \\
 Indonesian mango leaves & Mango leaves affected by pest & 510 \\
\bottomrule
\end{tabular*}
\end{table}

Some other proprietary mango leaf datasets are used in a few research works (detailed in Section~\ref{sec:related work}). However, these works deal with only a single type of disease, whereas we are interested in many diseases at the same time. Also, the size of those datasets is small.

So from our survey of existing literature it is evident that there is no standard dataset of images of mango leaves. Hence we think that it is imperative to develop such a dataset and release it to foster research in machine learning-based plant disease detection. Top machine learning scientists and practitioners often believe that the benefit of this great discipline is not yet fully harnessed for social good such as healthcare and agriculture.\footnote{\url{https://fortune.com/2019/09/10/a-i-s-next-big-breakthrough-eye-on-a-i/}} % \url{https://fortune.com/2021/07/30/ai-adoption-big-data-andrew-ng-consumer-internet/}}.
So our venture to prepare  a standard agricultural dataset will leap forward, however small, the endeavour of sharing the benefit of machine learning for mass people.

\subsection{Contribution}
The contributions of this research are summarized below:
\begin{itemize}
    \item We develop the first-ever dataset of mango leaf images of Bangladesh, one of the top mango-growing countries of the world. All the 1800 images are manually captured by camera from various mango orchards and then labelled by human experts. Further, after zooming and rotating some images, the size of the dataset reaches 4000.
    \item We cover a large number of major diseases (seven in particular) that attack the mango trees.
    \item We apply various data validation techniques that transform the raw dataset into a processed one.
    \item The dataset is released for public use in Mendeley Data Repository\footnote{\url{https://data.mendeley.com/}} and is readily available for downloading so researchers can fit the data directly into machine learning systems.\footnote{\url{doi: 10.17632/hxsnvwty3r.1}}
    \item As benchmark figures, we report prediction results of three machine learning models that are fit on the dataset.
    \item Although the dataset contains images of mango leaves of Bangladesh only, considering its large size, this can easily be used in a transfer learning setting \cite{ibrahim_torrey2010transfer} to predict mango diseases of other countries as well.
\end{itemize}
%According to United Nations, the world's current population is 7.6 billion which will reach 9.8 billion by 2050 \cite{8529755}.
The rest of the paper is organized as follows. Section~\ref{sec:method} explains in detail the steps of dataset preparation workflow. It also discusses the salient properties of the dataset. Section~\ref{sec:models} describes the machine learning models we apply as benchmark performance. Section~\ref{sec:related work} discusses the relevant existing research.  Section~\ref{sec:conclusion} concludes the paper with hints for future research directions.

\section{Methodology}
\label{sec:method}
The role of data is tremendously important in machine learning to the extent that it is believed by the practitioners that it is the quality and quantity of the data, and not the mathematical model, that plays the pivotal role in performance of modern machine learning systems.\footnote{Here the assumption is that a stat-of-the-art machine learning model is used. In technical terms, this philosophy is called the data-centric, as opposed to model-centric, approach \cite{ibrahim_url4}.} That is why researchers must follow the standard practices from the beginning to the end of dataset preparation procedure. In this section, firstly, we describe the steps we take in our dataset preparation task. Secondly, we analyze the visual characteristics of leaf images pertaining to different diseases. Thirdly, we list the key challenges we faced during the dataset development.

\subsection{Steps of Dataset Collection and Preparation}

%The whole dataset has only mango leaf images manually captured by ourselves. The images were collected in 2021 from four mango orchards of Bangladesh.

The main phases of our entire dataset preparation procedure are as follows:
\begin{enumerate}
    \item Conducting background study on prevalent diseases that affect mango trees.
    \item Selecting the mango orchards for data collection in consultation with the agricultural experts.
    \item Physically capturing the images of healthy and diseased mango leaves  from the trees. We consider seven diseases in total.
    \item Validating the images of the dataset. This step includes:
    \begin{itemize}
    \item Labelling the images manually by human experts.
     \item Resizing the images to standard shape.
     \item Cleaning the images from background noise.
    \end{itemize}
\end{enumerate}

We also apply some benchmark machine learning models on the dataset to see how well these models can leverage the newly-formed dataset. These results are discussed in Section~\ref{sec:method}.

Below we elaborate each of the above-mentioned four steps:

\subsubsection{Studying the Common Diseases of Mango Leaves}
Various diseases of mango trees greatly affect the yield. Many of these diseases are manifested in the leaves of a tree. Common such diseases include Dieback, Powdery Mildew, Red Rust, Cutting Weevil, Bacterial Canker, Sooty Mould, Anthracnose, Gall Midges, etc. Studies \cite{mia2020} show that about 39\% of mango trees are affected by Anthracnose whereas Powdery Mildew damages up to 23\% of unsprayed trees. Bacterial Canker, a deadly disease that can damage mango yields by 10\% to 100\% \citep{mia2020}. These figures give us a glimpse of the danger of not detecting diseases at an early stage.

The main diseases of mango trees that expose themselves in the leaves are briefly discuss below.

Anthracnose is produced by the fungus \emph{Colletotrichum gloeosporioides} and is considered to be the most devastating disease \cite{ibrahim_anthracose}.  %\cite{GreenLife}.
Bacterial Canker is another deadly mango disease caused by \emph{Xanthomonas axonopodis pv. mangiferaeindicae} \cite{Jonathan}, \cite{10.2307/43213883}. %which presumably originated in India and spread to other countries via the transfer of contaminated plant material .
%An insect in the beetle family Attelabidae is known as Deporaus marginatus, popularly referred to as the mango leaf Cutting Weevil.
The mango leaf Cutting Weevil is a destructive insect that attacks newly emerging mango foliage \cite{articleRashid}. Another significant disease is known as Die Back which is caused by the fungus \emph{Lasiodiplodia theobromae} \cite{articleKamil}, \cite{saeed2017detection}.
%The mango Gall Midge is a species of very small fly that can be found in almost all mango growing regions of the world.
The larvae of a very small fly called gall midges feed within the plant tissue, which results in an abnormal growth of the plant known. These galls can cause damage to the mango plant's leaves, flowers, fruit, and shoots \cite{ibrahim_gall}.  %\cite{BusinessQueensland}.
Powdery Mildew disease is caused by the fungus Oidium mangiferae, which is a plant pathogen that infects mango plants \cite{ibrahim_powdery}. %\cite{plantwise}. %The ascomycete disease known as powdery mildew of mango was first reported in 1914 by Berthet using samples that were gathered in Brazil. It is a member of the family Erysiphales and belongs to the genus Erysiphales \cite{Wikipedia}.
Sooty Mould, also known as Meliola Mangiferae, is one of the fungi that thrive on honeydew produced by sap-feeding insects. %The fungi cause just superficial damage to the tree and do not actually infect the plant tissue in any way. But because
This fungus blocks sunlight from entering into the chloroplasts in the leaf, thereby hurting the process of photosynthesis and the plant's growth \cite{Amir}.

\subsubsection{Selecting the Mango Orchards for Data Collection}

To collect data from mango gardens of different parts of the country, four mango gardens were selected based on their size and variety of trees. The selected orchards are: Sher-e-Bangla Agricultural University mango garden in Dhaka, Jahangir Nagar University garden in Savar, Udaypur village mango garden in Rajbari district, and Itakhola village mango garden in Nilphamari district. Our choice was proved to be correct as we found a good amount of diseased leaves from these gardens.

\subsubsection{Capturing the Leaf Images}

%As mentioned earlier, in this study we consider seven diseases, namely Anthracnose, Bacterial Canker, Cutting Weevil, Die Back, Gall Midge, Powdery Mildew, and Sooty Mould.
In addition to the seven diseases mentioned above, since the machine learning models need to recognize healthy leaf images as well as disease-affected ones, we include the healthy images as the 8th category in our dataset.

The leaf images are taken a few days before the winter of 2021. At first, the trees affected by the disease were located -- oftentimes with the help of agriculture experts. Then, the affected leaf images were taken out from the trees. Finally, some healthy leaf images were also taken. Some diseases were found in almost every tree whereas some other diseases were hard to find; for example, Bacterial Canker and Gall Midge disease-affected leaves were relatively low in number. After collecting the leaves, the images of all the leaves were captured individually using a camera with a white background. This way a total of around 6000 images are taken where eight categories of leaves are present.

\subsubsection{Validating the Dataset}

Since the image size of machine learning models must be of the same size, each of the captured images is resized to 240$\times$320 pixels and is saved in JPG format. The noises present in the images are manually cleaned, and in this process some severely hazy and noisy images are dropped out. This way a total of around 1800 images out of initial 6000 is retained. After that, zooming and rotation are performed on some images that resulted in 4000 images in total, where each of the eight categories has exactly 500 images. Note that there were a few leaves where traits of multiple diseases were present; we omit such leaves to reduce the noise in the dataset.\footnote{To be able to successfully classify unseen instances using a machine learning algorithm, there must be distinguishable patterns among various categories of the dataset.} Although we ourselves did study thoroughly the traits of leaves affected by different diseases,  sometimes we corroborated our judgement on labelling the images with agricultural experts. This makes the quality of labelling even more reliable.

\begin{figure}[htbp]
%\centerline{\includegraphics[width=80mm,scale=0.5]{figs/dataDiagram (4).jpg}}
\centerline{\includegraphics[width=60mm,scale=0.5]{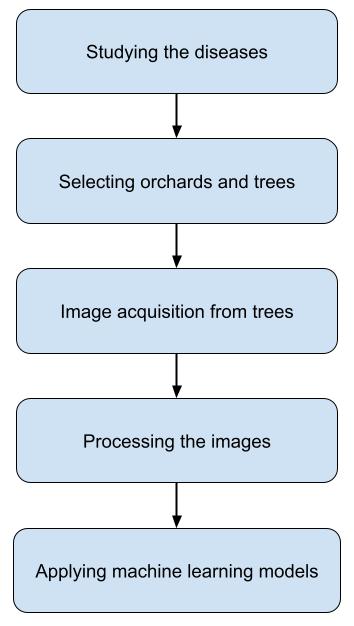}}
\caption{Flowchart showing the data preparation steps followed by model fitting.}
\label{fig:flowchard data collection}
\end{figure}

\begin{figure}[h!]
    \includegraphics[width=.24\textwidth]{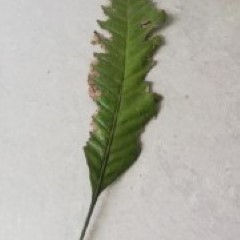}\hfill
    \includegraphics[width=.24\textwidth]{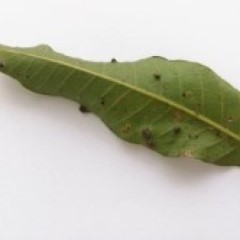}\hfill
    %\\[\smallskipamount]
    \includegraphics[width=.24\textwidth]{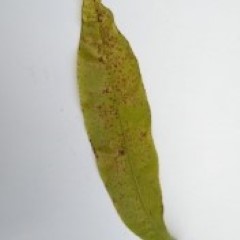}\hfill
    \includegraphics[width=.24\textwidth]{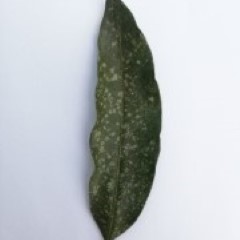}\hfill
    %new
     \includegraphics[width=.24\textwidth]{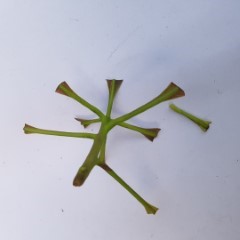}\hfill
    \includegraphics[width=.24\textwidth]{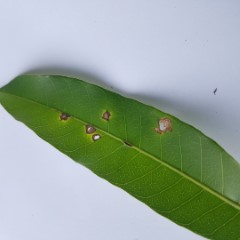}\hfill
     \includegraphics[width=.24\textwidth]{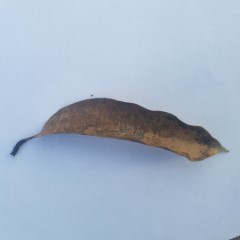}\hfill
    \includegraphics[width=.24\textwidth]{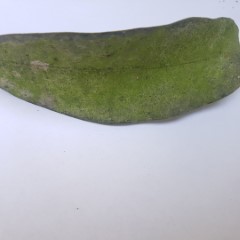}\hfill
    \caption{Sample raw images}\label{fig:foobar}
    \label{fig:sample raw images}
\end{figure}

\begin{figure}[h!]
    \includegraphics[width=.25\textwidth=]{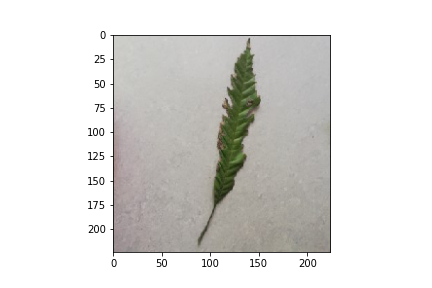}\hfill
    \includegraphics[width=.25\textwidth]{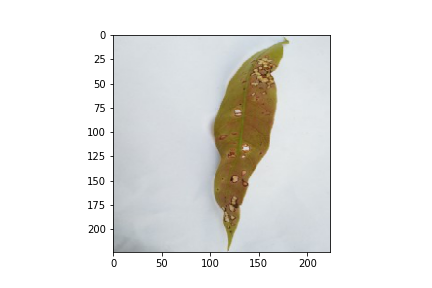}\hfill
   % \\[\smallskipamount]
    \includegraphics[width=.25\textwidth]{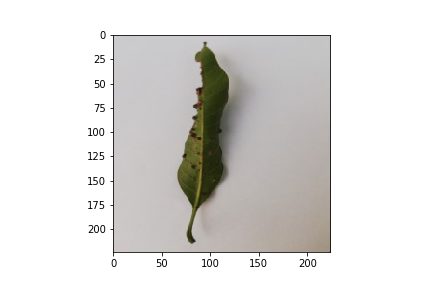}\hfill
    \includegraphics[width=.25\textwidth]{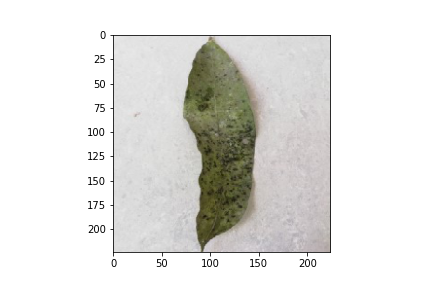}\hfill
    \caption{Sample processed images}\label{fig:foobar}
    \label{fig:sample processed images}
\end{figure}

The flowchart of data preparation stages is shown in Figure~\ref{fig:flowchard data collection}. Figures~\ref{fig:sample raw images} and \ref{fig:sample processed images} depict sample raw and processed images of the dataset respectively.  %Table~\ref{tbl1} displays the total number of images for each category.
Table~\ref{tbl2} depicts all the information of the dataset at a glance.

\begin{comment}
\begin{table}[width=.9\linewidth,cols=4,pos=h]
\caption{Details of image categories }\label{tbl1}
\begin{tabular*}{\tblwidth}{@{} LLLL@{} }
\toprule
Class name & Number of images \\
\midrule
 Anthracnose & 500 \\
  Bacterial Canker & 500\\
 Cutting Weevil & 500 \\
 Die Back & 500 \\
  Gall Midge & 500 \\
  Healthy & 500 \\
  Powdery Mildew & 500 \\
 Sooty Mould & 500 \\
\bottomrule
\end{tabular*}
\end{table}
\end{comment}

\begin{table}[width=.9\linewidth,cols=4,pos=h]
\caption{MangoLeafBD dataset information at a glance}\label{tbl2}
\begin{tabular*}{\tblwidth}{@{} LLLL@{} }
\toprule

 Type of data & 240x320 mango leaf images.\\
Data format & JPG. \\
Number of images & 4000 images. Of these, around 1800 are of distinct leaves, \\ & and the rest are prepared by zooming and rotating where deemed necessary. \\ %No data augmentation or replication techniques are applied here, \\& implying that all these 4000 images are distinct. \\
Diseases considered & Seven diseases, namely Anthracnose, Bacterial Canker, Cutting Weevil,  \\
  & Die Back, Gall Midge, Powdery Mildew, and Sooty Mould.\\
  Number of classes & Eight (including the healthy category). \\
Distribution of instances & Each of the eight categories contains 500 images. \\
How data are acquired & Captured from mango trees through mobile phone camera. \\
 %Experimental Factors  & All images are classified by diseased name \\
 Data source locations
  & Four mango orchards of Bangladesh, namely Sher-e-Bangla Agricultural University \\
   & orchard, Jahangir Nagar University orchard, Udaypur village mango orchard, \\ & and Itakhola village mango orchard. \\
  Where applicable & Suitable for distinguishing healthy and diseases leaves (two-class prediction)\\ &  as well as for differentiating among various diseases (multi-class prediction).\\
\bottomrule
\end{tabular*}
\end{table}

\subsection{Traits of Different Diseases in the Leaf Images}

\begin{figure}[htbp]
\centerline{\includegraphics[width=.24\textwidth]{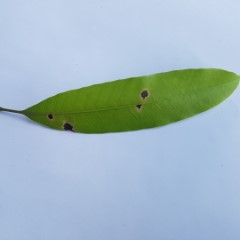}
\includegraphics[width=.24\textwidth]{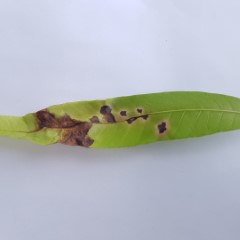}
}
\caption{Anthracnose}
\label{fig:antrax}
\end{figure}

\begin{figure}[htbp]
\centerline{\includegraphics[width=.24\textwidth]{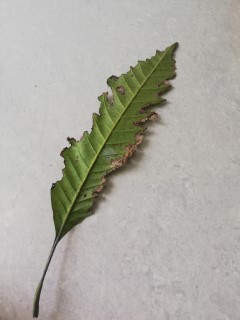}
\includegraphics[width=.24\textwidth]{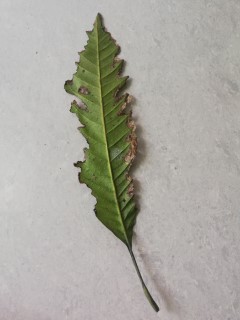}
}
\caption{Bacterial Canker}
\label{fig:bacteria}
\end{figure}

\begin{figure}[htbp]
\centerline{\includegraphics[width=.24\textwidth]{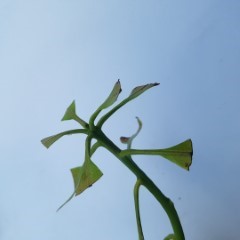}
\includegraphics[width=.24\textwidth]{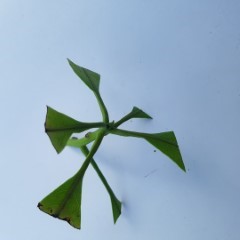}
}
\caption{Cutting Weevil}
\label{fig:cutting}
\end{figure}

\begin{figure}[htbp]
\centerline{\includegraphics[width=.24\textwidth]{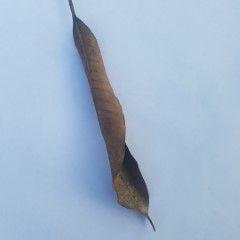}
\includegraphics[width=.24\textwidth]{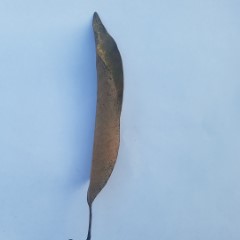}
}
\caption{Die Back}
\label{fig:dieback}
\end{figure}

\begin{figure}[htbp]
\centerline{\includegraphics[width=.24\textwidth]{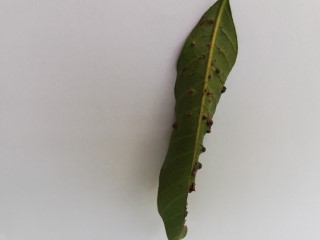}
\includegraphics[width=.24\textwidth]{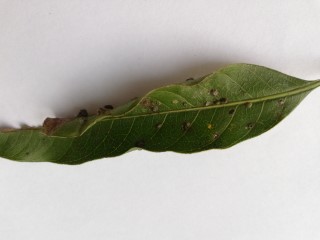}
}
\caption{Gall Midge}
\label{fig:gall}
\end{figure}

\begin{figure}[htbp]
\centerline{\includegraphics[width=.24\textwidth]{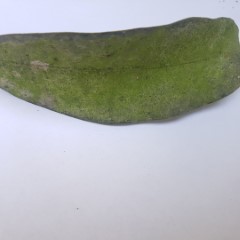}
\includegraphics[width=.24\textwidth]{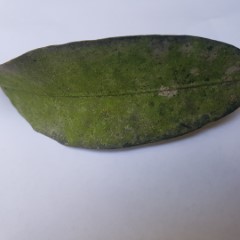}
}
\caption{Powdery Mildew}
\label{fig:powder}
\end{figure}

\begin{figure}[htbp]
\centerline{\includegraphics[width=.24\textwidth]{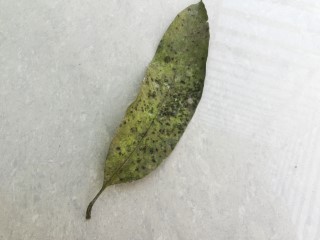}
\includegraphics[width=.24\textwidth]{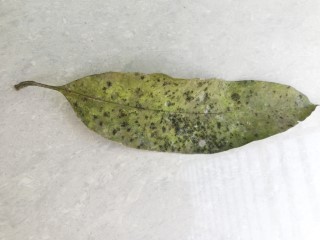}
}
\caption{Sooty Mould}
\label{fig:scooty}
\end{figure}

\begin{figure}[htbp]
\centerline{\includegraphics[width=.24\textwidth]{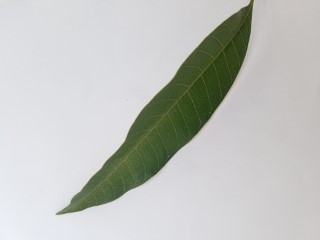}
\includegraphics[width=.24\textwidth]{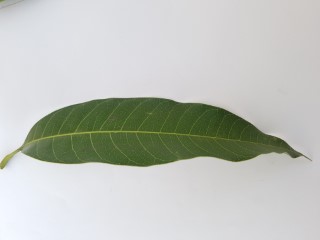}
}
\caption{Healthy}
\label{fig:healthy}
\end{figure}

As mentioned earlier, the instances of different classes pertaining to a dataset must have distinctive traits so that a machine learning model can effectively distinguish among the inter-class feature vectors during prediction phase. In this section we analyze the distinct traits of various diseases found in the leaf images of our dataset.  Figures~\ref{fig:antrax}, \ref{fig:bacteria}, \ref{fig:cutting}, \ref{fig:dieback}, \ref{fig:gall}, \ref{fig:powder}, \ref{fig:scooty}, and \ref{fig:healthy} show two sample images of each of the seven diseases and healthy categories.

%healthy

When Anthracnose is present, black necrotic patches emerge on both sides of the mango leaf. In most cases, necrotic patches form along the leaf margins where the lesions merge. Leaves that have been severely affected begin to curl. Young tissue is the primary site of infection, but conidia can be seen in lesions of all ages; see Figure.~\ref{fig:antrax}. When infected by Bacterial Canker, the bacterium pseudomonas mangifera causes mango fruits, leaves, stalks, and branches to get water-soaked spots that turned into typical cankers; see Figure~\ref{fig:bacteria}. Cutting Weevil disease cuts the mango leaf in such a way that it looks like it is cut with scissors; see Figure~\ref{fig:cutting}. Dieback is a disease that causes mango twigs to dry out and break off from the top down. This is followed by the leaves turning brown, drying out, and falling off; see Figure~\ref{fig:dieback}. Gall Midge disease causes leaves have what look like pimples on them. Heavy outbreaks of mango Gall Midge disease result in defoliation and reduced fruit yield; see Figure~\ref{fig:gall}. The white, powdery growths of fungus on the surface of leaves, flower stalks, flowers, and young fruits are a sign of the disease; see Figure~\ref{fig:powder}. Honeydew is a sticky, sweet secretion that some insects make to attract other insects. The sooty mould grows on honeydew. Using the honeydew as food, the mold slowly spreads over the surface of the affected plant part, turning it black in different ways; see Figure~\ref{fig:scooty}.

So we can see from the above discussion that there are  reasonably distinctive features among various disease classes of the images, thereby making it a fertile domain for applying machine learning models.

\subsection{Challenges to Prepare a Leaf Image Dataset}
Preparing a machine learning dataset from scratch is not a trivial task, rather it is a tedious one. It requires significant human resources and time. But eventually this effort, time, and labour pay off since a well-prepared dataset, if released for public use, is utilized by thousands of machine learning practitioners and researchers. Below we list the key challenges faced during our dataset preparation task:
\begin{itemize}
    \item Diversity of leaf diseases: as we discussed in Section~\ref{sec:method}, there is a large number of mango diseases.
    \item Geographical spread of mango orchards: researchers need to carefully choose the orchards and trees so that a good number of leaves of various diseases are found.
    \item Technical difficulties during image acquisition: mango trees are quite tall so it is difficult to get physically close to the leaves.
    \item Overwhelming choices of data validation techniques: deciding which methods will be useful and which are not requires decent background knowledge on data science and machine learning.
\end{itemize}

\section{Applied Machine Learning Models and Result Analysis}
\label{sec:models}
%It is important for other researchers  future researchers may use these figures as benchmark. So
In this section, we provide performance metrics of some machine learning models learnt from our dataset.

\subsection{Machine Learning Models}
We employ three deep learning models, namely  a CNN model, ResNet50 model, and a blend of SVM and CNN models (which we call CNN-SVM), and report their accuracy in terms of precision, recall and F1 measures. The technical details of these three models along with their parameter settings are given in Appendix~\ref{sec:appendix}.

\subsection{Result Analysis}

Accuracy is the percentage of correct predictions on the test set. Precision is the ratio of true positive cases to the predicted positive ones. Recall is the ratio of predicted true positive cases to all actual positive ones.\footnote{Precision and recall are measured on a class-by-class basis, whereas accuracy is measured considering all the classes.} Since precision and recall trade-off, i.e., neither precision nor recall alone can evaluate the overall goodness of a model, we need a combining metric; and F1 score serves this purpose which is the geometric mean of precision and recall.

%To do so, a confusion matrix was first created. The rows of the confusion matrix indicate the ground truth classes, while the columns provide information about DeepCONVSVM's predicted classes. Using this information, the accuracy and recall were computed.

%The proposed DeepCONVSVM's confusion matrix is shown in Fig. 14. The model has a Macro average precision of 90\% and a macro average recall of 91\%. Whereas both weighted average of precision and weighted average of recall have found of 91\%.
%final

\begin{table}[width=.7\linewidth,cols=4,pos=h]
\caption{Performance comparison of the three deep learning models}
\begin{tabular*}{\tblwidth}{@{} LLLL@{} }
\toprule
 Model & Precision & Recall &F1 Score \\
\midrule
  CNN & 87\%&85\%&85\%\\
 ResNet50 & 79\%&75\%&79\%\\
 CNN-SVM &91\%&90\%&90\%\\
\bottomrule
\end{tabular*}
\label{tab:accuracy}
\end{table}

In Table~\ref{tab:accuracy}, the performance of CNN, ResNet50 and CNN-SVM is shown. We see that all three models perform reseanably well with CNN-SVM model being the top in the list. %With 91\% accuracy, 91\%  precision, 90\%  recall, and 90\%  f1 score, the proposed architecture outscored the other two models.

%\begin{comment}  %FLAG: FOR DATA-IN-BRIEF ARTICLE. JUST REMOVE FOR OTHERS.

\section{Related Work}
\label{sec:related work}
In this section we discuss some works that use machine learning techniques to predict disease from plant diseases. The plants include tomato, pepper, potato, wheat, maize, rice, citrus etc. Very few papers deal with mango leaves, however, they work on proprietory datasets, i.e., datasets that are unavailable for research.

Mia et al. \cite{mia2020} work with a very small sample of only 20 proprietary images to classify four mango diseases. %the authors do not release their dataset for future use.
Saleem et al. \cite{ibrahim_saleem2021mango} use a small dataset consisting of less than 100 images collected in Pakistan for classifying only two diseases (Sooty Mould and Powedery Mildew) along with a healthy class. However, the authors do not share their dataset. Kumar et al. \cite{kumar2021classification} develop a CNN architecture based on VGG-16 to detect diseases in leaf images found in the PlantVillage dataset. The authors also work with some self-captured images of mango leaves affected by only Anthracnose disease. Singh et al. \cite{singh2019multilayer} present a multilayer convolutional neural network inspired by AlexNet architecture. They also mainly work with the PlantVillage dataset and some mango leaf images infected with the Antharacnose disease. %total 2200 images only with 1000 mango leaves?
Merchant et al. \cite{merchant2018mango} use unsupervised machine learning, specifically, clustering technique to categorize various nutrient deficiencies of mango leaves. The authors work on a small proprietary dataset collected from India.

Too et al. \cite{too2019comparative} use four deep CNNs, namely VGG-16, ResNet, DenseNet, and InceptionNet, to classify healthy and diseased leaf images of PlantVillage dataset. Gandhi et al. \cite{gandhi2018plant} use CNN and Generative Adversarial Networks (GAN) to identify the diseases of plant leaf images of PlantVillage dataset using a mobile application. %This work achieved 92\% accuracy.
Durmucs et al. \cite{durmucs2017disease} apply AlexNet and SqueezeNet models to classify plant leaf diseases in PlantVillage dataset.

Picon et al. \cite{picon2019deep} utilize DCNNs for the categorization of three fungal infections of wheat plant. The images are collected by the authors from two crop fields. Golhani et al. \cite{golhani2018review} employ various neural networks to classify plant images.  %With context to hyperspectral images, various imaging concepts are presented.
Iqbal et al. \cite{iqbal2018automated} identify the citrus plant leaf diseases. %The authors study various methodologies associated with detecting the disease. % and also have discuss the advantages and disadvantages of the studies.
Barbedo et al. \cite{barbedo2018factors} discuss different factors and challenges that affect the performance of a network. To show the factors and challenges of a network, a CNN model is used on plant images from DigiPath dataset that have twelve different diseases. Ma et al. \cite{ma2018recognition} use DCNN to identify four diseases in cucumber plants using a self-made dataset. % In this study, they tested four cucumber diseases and used real-time photos to verify the results. They propose to use thermal infrared imaging as an advancement in their work.
Zhangn et al. \cite{zhang2018identification} use a deep CNN model, namely GoogLeNet to detect diseases in maize plant. % The findings were confirmed using the PlantVillage dataset and images from Google.com. Incorporating mobile devices and other plant diseases is something that the authors are really excited about. In terms of accuracy, these models outperformed other transfer networks.
Lu et al. \cite{lu2017identification} present a DCNN model for the detection of ten diseases in rice plants. %They have worked with 10 healthy and diseased classes of 500 images and self-captured data is used in the validation process. %The 10-cross-fold validation approach is used to improve classification accuracy.
Jain et al. \cite{jain2017cloud} develop a CNN architecture to classify the diseases of firecracker and pomegranate plant leaves. % in real-time. This work is done in a cloud-based environment. The images were collected from a real-time environment.

%Convolutional neural networks and linear SVMs for image classification in \cite{DBLP:journals/corr/abs-1712-03541} have been proposed by Fred M. Agarap. They have worked with 10 classes of 70000 images and the validation is done on MNIST dataset.

%A deep learning algorithm based on linear support vector machines has been suggested by Yichuan Tang in \cite{tang2013deep}. On the common deep learning datasets MNIST, CIFAR-10, and the ICML, his results using L2-SVMs indicate that just replacing.

From the above discussion we see that although a good number of works have been performed by researchers on use of machine learning models to detect plant disease from leaf images, no work, to the best of our knowledge, investigates the mango leaf diseases using a standard, sizable, and publicly available dataset. The contribution of our research is expected to fill this gap in the literature.

%\end{comment}  %FLAG: FOR DATA-IN-BRIEF ARTICLE. JUST REMOVE FOR OTHERS.

\section{Conclusion}
\label{sec:conclusion}

The agriculture sector is yet to witness the benefit of machine learning discipline. One of the major barriers to utilize this powerful branch of applied science is the lack of representative data. In this work we aim to contribute to this domain by devising a standard, publicly available dataset of 4000 mango leaf images consisting of about 1800 distinct leaves. The images were manually captured by us from four mango orchards of Bangladesh,  cleaned from background noise, and resized to make the dataset ready-to-use by researchers and practitioners. We have also provided benchmark performance of some machine learning models. By utilizing these models,  diseases of mango trees can be detected relatively easily, economically, and at scale. The dataset has the potential to be further enlarged using various data augmentation techniques. We believe that the dataset should leap forward, however small, technology-based and automated agriculture.

%PREVIOUS:  Diseases need to be identified in the early state to increase mango production. To increase mango production, it is essential to identify diseases in their early state. Therefore, in this study, we have proposed a comparative model that identifies almost all kinds of common diseases in mango leaf with higher accuracy. For that, we have made our dataset that contains eight classes and then implement various models which are resnet50, CNN, and finally, CNN with SVM. Among these models we have found that DeepCONVSVM performs much better with an accuracy of 91\%. Further, this model can be used in real-time mobile app development which will accurately diagnose the disease in a very short time. As a result, this will stop the spread of diseases and improve the quality of fruit and increase productivity.

%part main

%\begin{comment}  %FLAG: FOR DATA-IN-BRIEF ARTICLE. JUST REMOVE FOR OTHERS.

\appendix
\section{Appendix: Technical Details of Applied Machine Learning Models}
\label{sec:appendix}
%Appendix sections are coded under \verb+\appendix+.

In this section, we briefly describe, in technical terms, the three machine learning models used in this research which is followed by a description of their parameter settings.

\subsection{Model 1: CNN}
As our first model, we employ Convolutional Neural Network (CNN) which is a special form of artificial neural network (ANN). ANN mimics  the neurons of human brain. A neuron is connected to another through exchange of information and finally makes a decision. It is commonly used for image and video processing. CNN architecture is a feed-forward network where neurons are attached to a small portion of the layers rather than being fully connected to all neurons, and that is why the total number of neurons is reduced~\cite{kumar2021classification}. %So even if the size of the image is larger, there is no need neurons too much, so the number of layer decreases.
CNN mainly consists of three types of different layers: convolution layer, pooling layer, and fully connected layer. The first one is the convolution layer and most of the time ReLU (Rectified Linear Unit) activation function is attached to it. Convolution layers extract different features from the input. %In these layers, it multiplied feature with the input image corresponding pixel then added and finally divide by the total number of the pixels in the feature.
Low-level features such as lines, edges, and corners are extracted by earlier convolution layers and later layers extract higher-level features \cite{hijazi2015using}. Pooling layers are placed in between convolution layers that reduces the resolution of the feature as well as the overfitting tendency. In pooling layers, 2$\times$2 or 3$\times$3 sized windows are usually selected, and for each window the maximum value or the average value may be picked. %Pooling may be done in two ways, maximum pooling and average pooling\cite{hijazi2015using}. If the maximum value is taken from the window, it is called maximum pooling and if it is taken average value from the window it is called average pooling.
The fully-connected layer is the last layer of a neural network that produces the final prediction of the input image.

In our architecture, we use five convolutional layers where the
first two layers have 64 filters, and the second two layers have 128 filters, and the the fifth layer has 256 filters. ReLU activtion function is used after each layer.  %Pooling layers reduce the resolution of the features. In pooling layers, the  2×2 or 3×3 window sizes are selected and for each window, the maximum value is picked. Pooling may be done in two ways, maximum pooling and average pooling [7]. If the maximum value is taken from the window, it is called maximum pooling and if it is taken average value from the window it is called average pooling.
We also apply three max pooling layers, and a fully-connected layer is the last layer of the network. Finally, in the dense layer 512 hidden neurons are used.

\subsection{Model 2: Resnet50}
Over the years, various specific architectures of CNN have gained momentum. As our second model, we employ such a special CNN called Residual Network or ResNet50 which is a relatively new CNN architecture proposed in 2015. It has 48 Convolution layers along with 1 maximum pool and 1 average pool layer. %ResNet is like another network in that it is made up of many layers, such as convolution, pooling, and fully-connected layers.
The ResNet50 model is split into five parts, each with its own convolution and identity block. Each convolution block has three convolution layers, and each identity block has three convolution layers. About 23 million parameters are trained in ResNet50. In a CNN, oftentimes the backpropagation error signal in a feed-forward network often reduces or increases exponentially depending on how far away it is from the last layer. This problem is called vanishing and exploding gradient~\cite{mukti2019transfer}.  %To solve this problem ResNet estimates the delta necessary to get from one layer to the next and achieve the final prediction.
ResNet solves the vanishing gradient problem by enabling gradient to flow along an additional shortcut path~\cite{theckedath2020detecting}. In our implementation of ResNet50, we use the Keras library of TensorFlow platform\footnote{\url{https://www.tensorflow.org/}} along with its default parameter settings.

\subsection{Model 3: Blend of CNN and SVM}
It is often found that using a traditional machine learning model as the last layer in a CNN yields better performance. Hence as our third model, we employ a blend of CNN and support vector machine (SVM). It is a clever algorithm that transfers the input vector to a feature space with a higher dimension. It finds the optimal hyperplane from two classes by maximizing the margin.

Our architecture combines CNN and SVM, with five convolution layers followed by the ReLU activation unit, three max-pooling layers, and two dense or fully-connected layers. As the output layer it uses SVM. %A flattened hidden layer is used to transform the photos into a 1D array. Convolutional and max-pooling layers each measure 3$\times$3 and 2$\times$2, but the size of the input images and feature maps changes. % as illustrated in Figure 4. %The layers of this model are as follows: a standard-sized image is transformed into a feature set for subsequent processing via a sequence of layers.
The first layer of this model is a convolution layer with 64 filters and ReLU as the activation function. The max pooling layer is the next layer, and it minimizes the size of the convoluted image to 2$\times$2. The following two layers are also convolution with 128 filters and ReLU activation units. A 2$\times$2 max pool layer follows. %There are two strides in the maximum pooling layer, which has a pool size of (2, 2).
Next is another convolutional layer with 256 filters and ReLU, followed by another convolutional layer with 128 filters and ReLU. A maximum pooling layer of size 2$
\times$2 follows it. After this,  flattening of the features occur that gives each image's final feature set as the output. This feature set is then fed into a fully connected layer having 512 neurons. Finally, in the output layer, an SVM is used to classify the image.

\subsection{Implementation Details}
%This study's classification objective\subsection{Model 1: CNN} necessitated a multi-class classification. Hence,

The three models are implemented using Python programming language in the TensorFlow framework using Google Colab platform and executed on a GPU.\footnote{TensorFlow is an open-source machine learning and deep learning library developed by Google. For details, please see \url{https://www.tensorflow.org/}.} %The entire training and validation process was carried out using free and open-source software. The model was created in Google Colab and executed on a GPU.

Since we deal with a multi-class classification problem, a categorical cross-entropy is utilized as the loss function in CNN and ResNet50 models. The squared hinge loss is employed for CNN-SVM model. Adam optimizer is used which is a stochastic optimization method for determining adaptive learning rates for parameters. %The trained models were validated with test images when the training was completed. Precision, Recall, F1-score, and Accuracy were used to evaluate the models' performance.
%In SVM, there is a parameter called  “kernel\_regularizer” and inside this regularizer, we have used l2 norm and pass softmax as activation function.
The performance of a model is dependent on the batch size. Models with larger batch sizes tend to perform better than those with smaller batch sizes. The three models are trained with a batch size of 128. Each model is trained for 100 epochs.

%Precision is determined by dividing the number of true positives by the total number of data points classified positively by the model.

Augmenting the images of the dataset using various computational techniques allows us to use different variations of the same image. To achieve this, we use TensorFlow image data generator that uses scaling, shearing zooming, flipping, among other operations. As such, the training phase of each of the algorithms work not with 4000 images, rather with much more. % because multiple images are created from a single image using the augmentation techniques. which includes the following operations: (1) scaling down by factor 255, (2) shearing by 10\%, (3) the parameter $Zomm\_range$ was set to 10\%, and (4) the parameter $horizontal\_flip$ was set true for mirror reflection.

%To achieve the required classification accuracy, a total of 100 epochs were used.

%Various image preprocessing was done prior to training.

%\end{comment}  %FLAG: FOR DATA-IN-BRIEF ARTICLE. JUST REMOVE FOR OTHERS.

\begin{comment}
\verb+\printcredits+ command is used after appendix sections to list
author credit taxonomy contribution roles tagged using \verb+\credit+
in frontmatter.

\printcredits
\end{comment}

%% Loading bibliography style file
\bibliographystyle{plain}
%\bibliographystyle{cas-model2-names}

% Loading bibliography database
\bibliography{cas-refs}

%\vskip3pt

\begin{comment}
\bio{}
Author biography without author photo.
Author biography. Author biography. Author biography.
Author biography. Author biography. Author biography.
Author biography. Author biography. Author biography.
Author biography. Author biography. Author biography.
Author biography. Author biography. Author biography.
Author biography. Author biography. Author biography.
Author biography. Author biography. Author biography.
Author biography. Author biography. Author biography.
Author biography. Author biography. Author biography.
\endbio

\bio{figs/pic1}
Author biography with author photo.
Author biography. Author biography. Author biography.
Author biography. Author biography. Author biography.
Author biography. Author biography. Author biography.
Author biography. Author biography. Author biography.
Author biography. Author biography. Author biography.
Author biography. Author biography. Author biography.
Author biography. Author biography. Author biography.
Author biography. Author biography. Author biography.
Author biography. Author biography. Author biography.
\endbio
\bio{figs/pic1}
Author biography with author photo.
Author biography. Author biography. Author biography.
Author biography. Author biography. Author biography.
Author biography. Author biography. Author biography.
Author biography. Author biography. Author biography.
\endbio
\end{comment}

\end{document}